\documentclass[conference]{IEEEtran}
\IEEEoverridecommandlockouts
\usepackage{cite}
\usepackage{amsmath,amssymb,amsfonts}
\usepackage{algorithmic}
\usepackage{graphicx}
\usepackage{subfigure}
\usepackage{booktabs}
\usepackage{textcomp}
\usepackage{xcolor}
\usepackage{colortbl}
\usepackage{makecell}
\usepackage{xspace}
\usepackage{caption}
\newcommand{\alias}{MGRAL\xspace}
\newcommand{\etal}{\textit{et al.\xspace}}
\definecolor{bestcolor}{gray}{.92}
\newcommand{\bestcell}[1]{\cellcolor{bestcolor}{#1}}
\def\BibTeX{{\rm B\kern-.05em{\sc i\kern-.025em b}\kern-.08em
    T\kern-.1667em\lower.7ex\hbox{E}\kern-.125emX}}
\begin{document}

\title{Aligning Data Selection with Performance: Performance-driven Reinforcement Learning for Active Learning in Object Detection}

\author{Zhixuan Liang$^{1,2}$\quad Xingyu Zeng$^{2}$\quad Rui Zhao$^{2}$\quad Ping Luo$^{1}$\\ ${}^{1}$The University of Hong Kong\quad ${}^{2}$SenseTime}

\maketitle

\begin{abstract}
Active learning strategies aim to train high-performance models with minimal labeled data by selecting the most informative instances for labeling. However, existing methods for assessing data informativeness often fail to align directly with task model performance metrics, such as mean average precision (mAP) in object detection. This paper introduces Mean-AP Guided Reinforced Active Learning for Object Detection (\alias), a novel approach that leverages the concept of expected model output changes as informativeness for deep detection networks, directly optimizing the sampling strategy using mAP.
\alias employs a reinforcement learning agent based on LSTM architecture to efficiently navigate the combinatorial challenge of batch sample selection and the non-differentiable nature between performance and selected batches. The agent optimizes selection using policy gradient with mAP improvement as the reward signal. To address the computational intensity of mAP estimation with unlabeled samples, we implement fast look-up tables, ensuring real-world feasibility.
We evaluate \alias on PASCAL VOC and MS COCO benchmarks across various backbone architectures. Our approach demonstrates strong performance, establishing a new paradigm in reinforcement learning-based active learning for object detection.
\end{abstract}


\section{Introduction}
\label{sec:intro}
The pursuit of artificial intelligence fundamentally revolves around optimizing two key elements: data and models. While significant advancements have been made in refining model architectures, the focus of contemporary research is increasingly shifting towards more efficient data utilization strategies. Among these, Active Learning (AL) stands out for its ability to efficiently train high-performance models with minimal labeled data. This approach is particularly valuable in environments where there is a continuous flow of operational data requiring high labeling costs. By strategically selecting and annotating the most informative ones, AL optimizes data utilization, significantly boosting the efficiency of AI systems.

Determining the most informative data typically involves identifying data points that are complementary to the currently labeled ones within the model's feature space, primarily characterized by uncertainty. Although various definitions of uncertainty have been proposed and utilized as query strategies in active learning~\cite{LL4AL,CDAL,MIAL,zhang2021learning,EBAL}, these do not always correlate directly with the performance metrics of the task model, such as mean Average Precision (mAP) in object detection. EMOC~\cite{freytag2014selecting} proposed the expected model output change to measure sample informativeness, but this approach was limited to Gaussian Process Regression models and lacks applicability to deep learning architectures. Most existing methods, including EMOC, evaluate samples individually, overlooking the collective impact of sample batches in AL scenarios.

\begin{figure}[tbp]
\centerline{\includegraphics[width=.99\columnwidth]{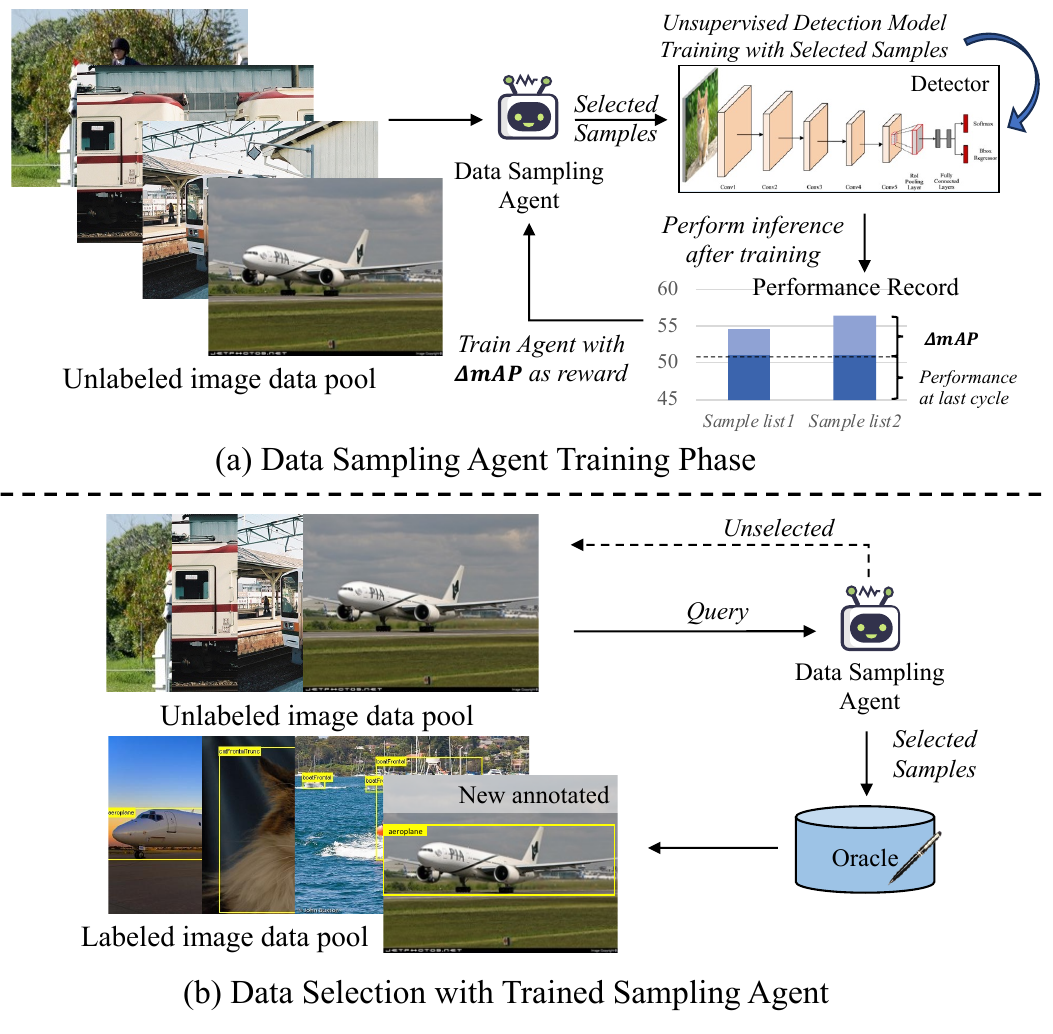}}
\vspace{-3pt}
\caption{\textbf{Training and selection pipelines of Mean-AP Guided Reinforced Active Learning~(\alias) for object detection.} (a) Data sampling agent training phase: The RL-based agent learns to select informative samples using $\Delta$mAP as reward, where performance gains are efficiently estimated through a semi-supervised detector. (b) Active selection phase: The trained agent directly processes the unlabeled pool to select samples for oracle annotation, which are then added to the labeled pool for detector improvement.}
\vspace{-10pt}
\label{fig:pipeline}
\end{figure}

In this paper, we introduce Mean-AP Guided Reinforced Active Learning (\alias), a novel approach that directly leverages mAP to guide data sampling for object detection. The core challenge lies in the discrete nature of batch selection from the unlabeled pool, making the relationship between selected samples and mAP changes non-differentiable. To address this, we employ a reinforcement learning agent that uses mAP variation ($\Delta$mAP) as the reward, optimizing the selection process through policy gradient techniques. This approach enables efficient exploration of possible batch combinations, maximizing mAP improvement per selected batch.

To estimate the potential change on mAP when not having labels of selected samples, \alias employs unsupervised model to approximate. This approach is effective as it captures global trends sufficient for identifying potentially impactful data. The unsupervised model's output need not be absolutely correct labels; rather, it aims to correctly estimate which data significantly differ from the model's current decision boundaries or distributions. This relative information is crucial for guiding the active learning process in selecting the most informative samples for labeling.
Additionally, to address the extensive training time required by the reinforcement learning agent, which necessitates retraining the semi-supervised model iteratively, we implement fast lookup tables for acceleration. Empirical results demonstrate our method's efficiency on both Pascal VOC~\cite{PASCALVOC} and MS COCO~\cite{COCO} datasets. 

Our contributions can be summarized as:
(1) We propose MAGRAL, a novel approach that extends the concept of expected model output changes to deep neural networks for object detection, directly aligning the sampling strategy with mAP.
(2) We introduce a reinforcement learning agent for efficient batch sample selection, addressing the combinatorial explosion and non-differentiable challenge and optimizing for mAP improvement.
(3) We implement practical techniques, including an unsupervised model for mAP estimation and a fast look-up table, making our method feasible for real-world active learning scenarios in object detection.

\section{Related Works}
\subsection{Performance-Driven Active Learning Strategies}
Classical active learning methods typically fall into uncertainty-based, distribution-based, and hybrid categories. Uncertainty-based methods (e.g., information theoretic heuristics~\cite{li2013adaptive}, query-by-committee~\cite{dagan1995committee,freund1992information}, and Bayesian models~\cite{golovin2010near,gal2017deep}) prioritize perplexing data points but may lead to redundancy. Distribution-based approaches like Core-set~\cite{coreset} aim for diversity but struggle with high-dimensional spaces~\cite{jain2016active,donoho2000high}. Hybrid methods~\cite{osugi2005balancing,li2013adaptive,kuo2018cost,yang2017suggestive,EBAL} attempt to balance both, yet face challenges in effectively combining these metrics. However, these methods may not directly align with the task model's performance.

Freytag~\etal~\cite{freytag2014selecting} proposed expected model output changes to measure sample informativeness, but it was limited to Gaussian process regression models and lacks applicability to deep learning architectures. Most previous methods focus on individual sample selection, neglecting batch selection importance. Our work extends these concepts to deep neural networks, directly aligning sampling with mAP for object detection and addressing batch selection challenges.

\subsection{Deep Learning and Reinforcement Learning in AL}
Deep learning has introduced methods that calculate informativeness by learning or use neural network-driven selection, such as learning loss estimation~\cite{LL4AL} and adversarial-based VAAL~\cite{VAAL}. Meanwhile, reinforcement learning has been adopted to learn better query strategies, using techniques like policy gradient methods~\cite{bachman2017learning}, imitation learning~\cite{liu2018learning}, bi-directional RNNs~\cite{contardo2017meta}, and Deep Q-Networks~\cite{konyushkova2018discovering,DBLP:conf/iclr/CasanovaPRP20}. However, these methods often struggle with batch sample selection and integrating multi-instance uncertainty within a single image for object detection tasks. Our approach addresses these challenges by using a reinforcement learning agent optimized for batch selection with $\Delta \text{mAP}$ as the reward.


\subsection{Active Learning for Object Detection}
Recent advances in active learning for object detection include LL4AL~\cite{LL4AL}, which adapts instance loss predictions, and Aghdam~\etal~\cite{aghdam2019active}, which combines uncertainty metrics for foreground objects and background pixels. CDAL~\cite{CDAL} enhances sample representativeness through spatial context, while MIAL~\cite{MIAL,wan2023multiple} employs adversarial classifiers and a semi-supervised framework. EBAL~\cite{EBAL} integrates uncertainty and diversity but faces challenges with computational complexity and class imbalance. Our method differs by directly utilizing mAP to guide the selection process, addressing the limitations of previous approaches in balancing various metrics and handling batch selection efficiently.

\section{Methodology}
\subsection{Problem Definition}
Active learning for object detection follows the setting that a small labeled set $X_L$ containing images with instance labels, denoted as $\{(x_L, y_L)\}$ and a large unlabeled set $X_U$ without labels, denoted as $\{(x_U)\}$ are given. The labels include locations of bounding boxes and their corresponding categories.

For each cycle of the active learning process, a detection model $M_i$ ($i$ denotes the cycle subscript) is initially trained using the labeled dataset $X_L^i$. Subsequently, active learning employs a query strategy to select a subset of images $X_S^i = \{(x_S^i)\}$ from the unlabeled dataset $X_U^i$. These images are then annotated and integrated into $X_L^i$ to create an updated labeled dataset $X_L^{i+1} = X_L^{i} \cup \{(x_S^i, y_S^i)\}$. The updated dataset $X_L^{i+1}$ is used for training the next iteration of the detection model $M_{i+1}$. This cycle repeats until the size of the labeled dataset reaches the predefined budget $B$. The effectiveness of the query strategy is pivotal, as it directly influences the enhancement of model performance with each cycle, motivating the development of our proposed method.

\subsection{Overview of~\alias Pipeline}
Our proposed method \alias introduces a novel approach to active learning for object detection that directly optimizes mean Average Precision (mAP) through reinforcement learning. The core innovation lies in integrating a reinforcement learning-based sampling agent into the conventional pool-based active learning pipeline, which learns to select the most informative samples by directly optimizing for mAP improvement.

As shown in Figure~\ref{fig:pipeline}, during training, \alias operates as a nested optimization process. The outer loop trains the sampling agent iteratively, while the inner loop focuses on calculating mAP improvement. Specifically, for each active learning round, (Inner Loop) we first construct a lookup table by parallel training multiple detector variants (Section~\ref{subsec:acceleration}). This table serves as an efficient approximation mechanism for estimating performance gains. (Outer Loop) The sampling agent, detailed in Section \ref{subsec:agent_architect}, then undergoes training iterations where it selects candidate sets from the unlabeled pool and receives feedback based on the estimated mAP improvements (Section \ref{subsec:reward_design}). This feedback drives the policy gradient updates of the sampling agent through our stabilized optimization process (Section \ref{subsec:training}).

During inference (active sample selection), the trained sampling agent directly processes the unlabeled pool through its LSTM-based architecture. For each image, it computes a selection score based on both the image's features and the historical selection context. The system then selects the top-scoring images within the specified budget for human annotation.

This design allows \alias to effectively bridge the gap between sample selection and model performance improvement, while maintaining computational efficiency through our lookup table acceleration technique. The following sections detail each component of this framework and their interactions.

\begin{figure}[t]
    \includegraphics[width=0.95\columnwidth]{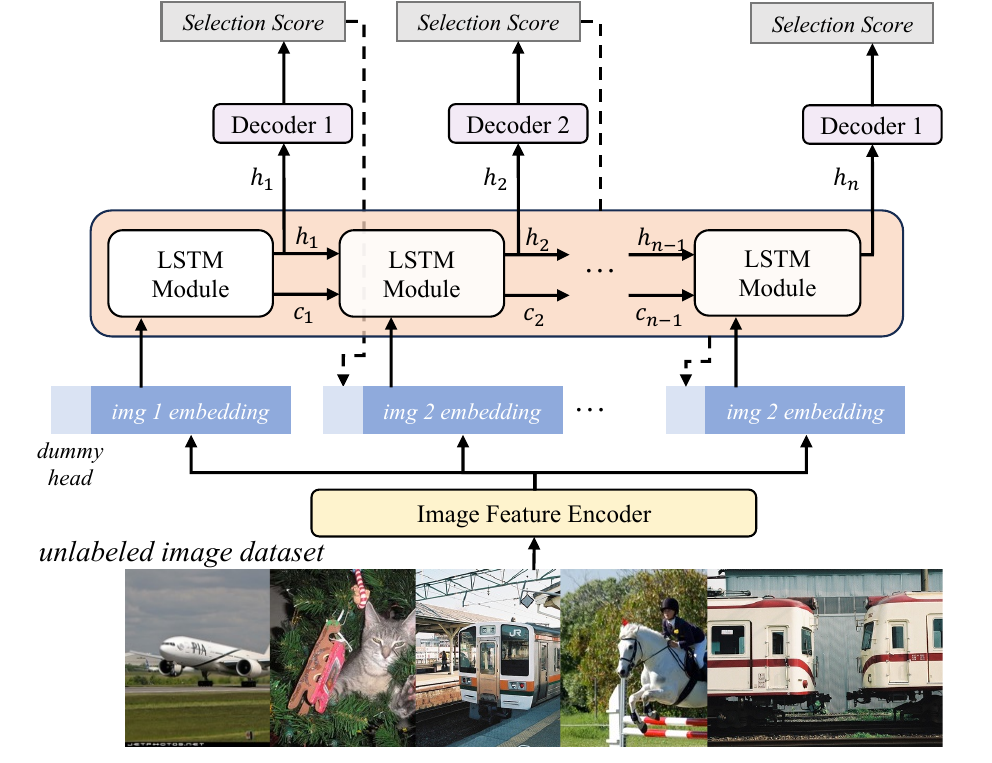}
    \vspace{-3pt}
    \caption{\textbf{Data sampling agent architecture.} The agent processes unlabeled images sequentially through three main components: (1) an Image Feature Encoder that extracts visual representations from input images, (2) a series of parameter-shared LSTM modules that process image embeddings while maintaining temporal dependencies through hidden states ($h_i$) and cell states ($c_i$), and (3) individual decoder networks that output selection scores for each image. A dummy head is used to initialize the first LSTM state. Each image embedding combines the feature representation with the previous decision, enabling considering both visual content and selection history.}
	\label{fig:controller}
    \vspace{-8pt}
\end{figure}

\subsection{\alias Data Sampling Agent}
\label{subsec:agent_architect}
The sampling agent adopts a Neural Architecture Search (NAS) inspired design, utilizing a Long Short-Term Memory (LSTM)~\cite{hochreiter1997long} network for sequential sample selection. As shown in Figure~\ref{fig:controller}, each image is first processed through a feature extraction network $\Phi$, which derives a feature vector using a pre-trained detection model. Then the extracted features then flow through parameter-shared LSTM modules that ensure scalability while preventing gradient vanishing. The final component is a 2-layer MLP decoder that outputs selection scores for each sample.

The agent's architecture integrates both current features and historical decisions through combining each image's embedding vector with the decision vector from the preceding unit, enabling the agent to consider both new data characteristics and previous selection patterns. This interaction is formally expressed as:
\begin{align}
\vspace{-5pt}
h_i, c_i &= \text{LSTM-Module}([\Phi(I_i), A_{i-1}], h_{i-1}, c_{i-1}),\\
A_i &= \Psi_i(h_i),
\vspace{-5pt}
\end{align}
where $[\cdot]$ denotes concatenation operation, $\Psi_i$ is the decoding network for the $i$-th image. 
During operation, the agent uses these computed scores to select the top-$B$ samples from the unlabeled pool, where $B$ is the active learning budget. This selection mechanism forms the basis for the reward computation and policy gradient updates detailed in Sec.~\ref{subsec:training}.

\subsection{Performance-Driven Reward Design}
\label{subsec:reward_design}
The reward mechanism forms the cornerstone of our approach by directly linking sample selection to detector performance through mAP improvement ($\Delta \text{mAP}$). During each training iteration i, the system combines the current labeled dataset $X_L$ with newly selected samples $X_S^i$. A new detector is trained on this combined set to compute $\text{mAP}_i$, with the reward calculated as $\Delta \text{mAP} = \text{mAP}_i - \text{mAP}_{i-1}$.

To address the challenge of evaluating unlabeled samples during the selection process (sampling agent training \& inference), we employ a semi-supervised detection model as a proxy for performance estimation. Specifically, during active selection of unlabeled data, we need to predict the performance gain that would result from oracle labeling, without having access to the actual annotations. We achieve this by using a semi-supervised model trained on $X_{i-1} + \text{(unlabeled)} X_B$ to approximate the performance gain of a supervised model trained on $X_{i-1} + \text{(labeled)} X_B$, where $X_B$ represents the candidate unlabeled batch.

Importantly, we utilize the semi-supervised model solely as a mechanism to capture global trends in the unlabeled data distribution and provide relative information about data informativeness. This approach enables us to rank and identify potentially impactful samples without requiring perfect predictions or pseudo-labels, making it an effective proxy for estimating the potential value of unlabeled samples once they are properly annotated.

\subsection{Training Process and Optimization}
\label{subsec:training}
The training process implements a robust optimization strategy centered on iterative refinement of the data sampling agent. For each iteration, the agent produces selection scores for unlabeled samples through its LSTM architecture and selects the top-$B$ scoring samples to form a binary mask. Using \textbf{policy gradient}, we enable reward gradients to flow back through this discrete selection process by multiplying the mask with the downstream reward gain, allowing effective update of the agent's parameters.

To stabilize the policy gradient training, we employ a moving average baseline mechanism that adapts to the changing performance landscape. Let $\text{mAP}_i$ denote the mean Average Precision achieved by the detector after incorporating the i-th batch of selected samples. We maintain a reference baseline $\text{mAP}_{ref}$ that tracks the average performance trend:
\begin{equation}
\text{mAP}_{ref} = \lambda * \text{mAP}_{ref} + (1 - \lambda) * (\text{mAP}_i -\text{mAP}_{ref}).
\end{equation}
where $\lambda$ is a momentum coefficient that controls the update rate of the reference baseline. This dynamic baseline helps normalize the reward signal and reduce variance in the policy gradient updates.

The agent's loss is then defined relative to this adaptive baseline:
\begin{equation}
loss_\text{agent} = -(\text{mAP}_i -\text{mAP}_{ref}).
\end{equation}

Here, $\text{mAP}_i$ represents the estimated performance after incorporating the current batch selection, while $\text{mAP}_{ref}$ serves as a moving average of historical performance values. By subtracting the baseline from the current performance, we obtain a more stable learning signal that indicates whether the current batch selection is better or worse than the recent average. This normalized reward, combined with the policy gradient approach, enables effective optimization of the sampling agent's parameters through the discrete selection process.

The training process continues for a fixed number of iterations or until convergence, with the agent progressively learning to identify more informative samples based on the relative performance improvements they provide. This policy gradient optimization framework allows the agent to learn effective sample selection strategies despite the discrete nature of the selection process and the non-differentiable mAP metric.

\subsection{Acceleration Technique}
\label{subsec:acceleration}
A critical challenge in our approach lies in the computational cost of the inner loop optimization. Without acceleration, each iteration of the sampling agent training would require a complete retraining of the semi-supervised detector to estimate mAP for every potential batch selection. This would result in an prohibitive computational overhead of 2200 (VOC) or 800 (COCO) \textbf{sequential} detector training iterations for each round of active learning.

To address this challenge, we introduce an efficient lookup table-based acceleration strategy. The system constructs this table by training $M$ semi-supervised models \textbf{in parallel}, each utilizing the current labeled data ($X_{labeled}$) plus a randomly selected batch of unlabeled data ($\widetilde{X}_B$) within the active learning budget $B$. This parallel pre-computation serves as a basis for rapid performance estimation during the sampling agent's training iterations.

During the agent training, instead of retraining the detector, performance estimation leverages the Wasserstein distance~\cite{vallender1974calculation} (one of optimal transport distance) to measure similarity between the visual representations of selected batches and pre-computed datasets. The system computes weights inversely proportional to measured distances, enabling accurate mAP approximation through weighted summation of the most similar pre-computed results. In cases where similarity measures exceed a predetermined threshold (mean feature distance minus one standard deviation), the system falls back to direct training for precise mAP calculation.

This acceleration approach dramatically reduces computational overhead, achieving a $1000\times$ speedup by transforming the requirement from thousands of sequential detector training iterations to a single parallel training phase per round of data selection. This efficiency gain makes the proposed method practically feasible while maintaining its effectiveness in selecting informative samples.


\begin{figure*}[t] \centering
\begin{minipage}{0.6\textwidth}
\vspace{-5pt}
\subfigure[SSD on Pascal VOC] {
\label{fig:voc_result}   
\includegraphics[width=0.48\textwidth]{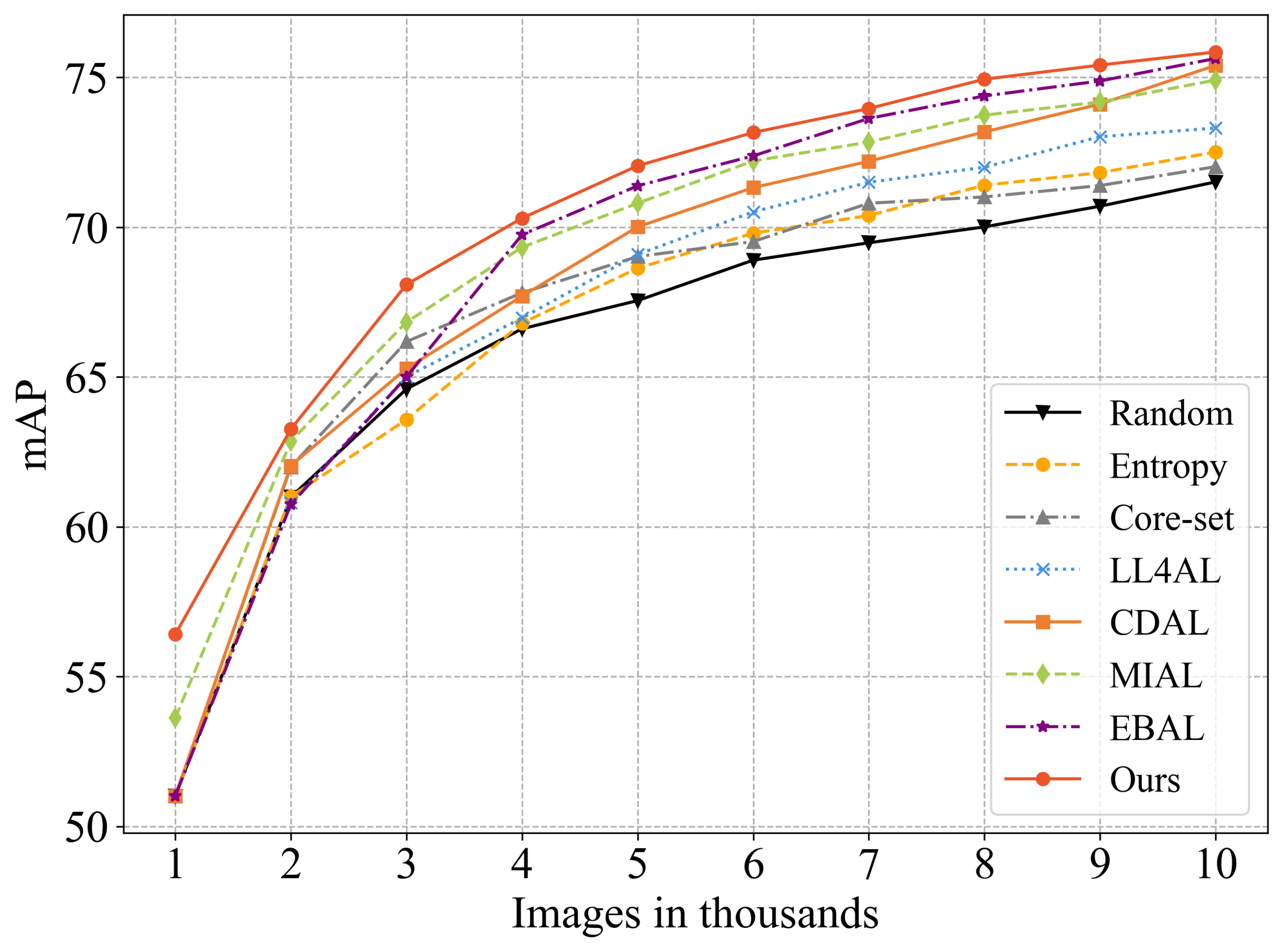}  
}
\hspace{-8pt}
\subfigure[RetinaNet on MS COCO] { 
\label{fig:coco_result} 
\includegraphics[width=0.48\textwidth]{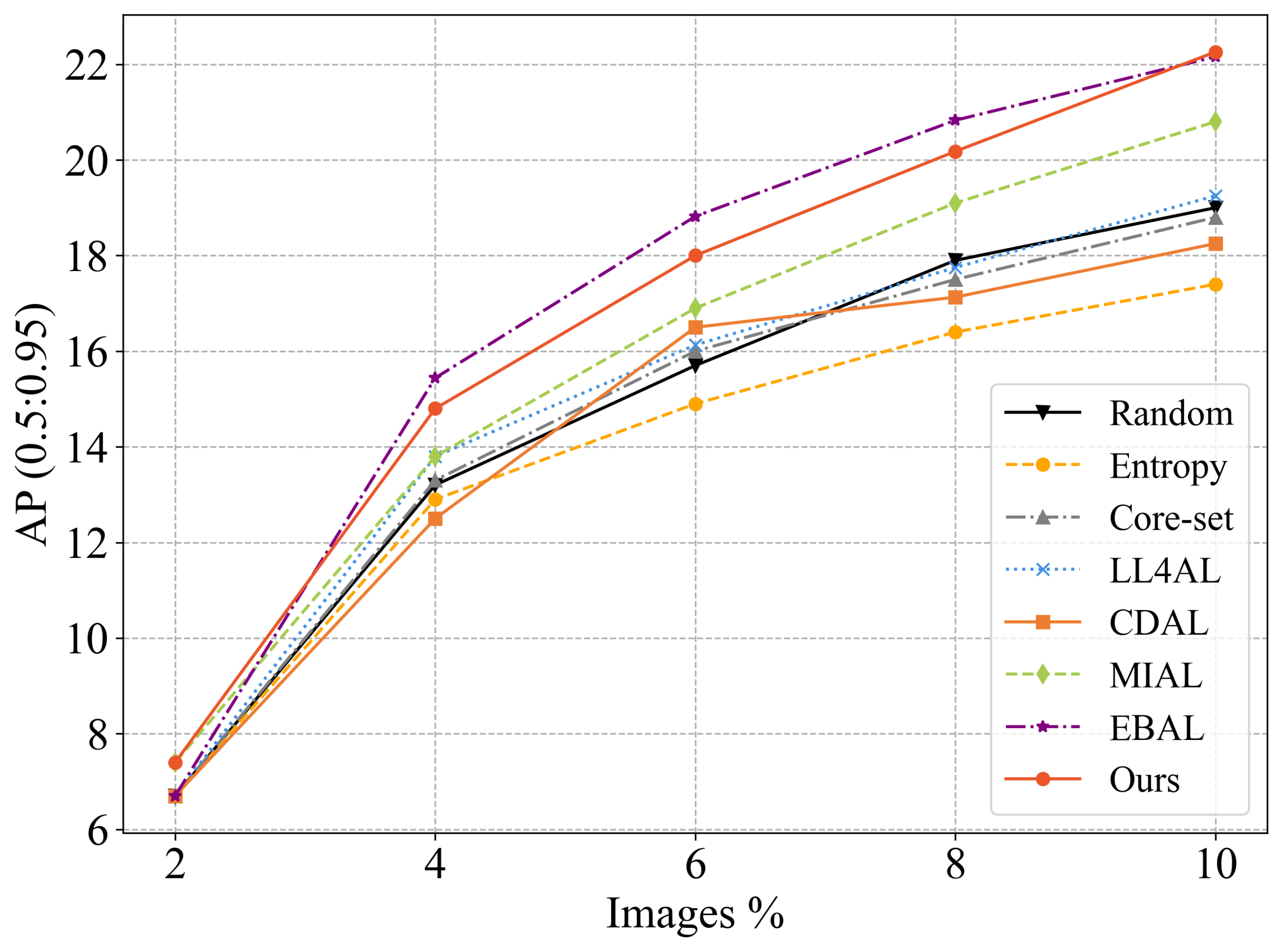}  
}
\vspace{-8pt}
\caption{\textbf{Comparative performance of active learning methods.}} 
\label{fig:overall_result}
\vspace{-10pt}
\end{minipage}
\begin{minipage}{0.39\textwidth}
\centering
  \small
  \captionof{table}{\textbf{\textsc{Training time of \alias with and without lookup table on VOC.}}}
  \vspace{-5pt}
  \resizebox{\textwidth}{!}{
  \begin{tabular}{c|c|c}
    \toprule
    \textbf{Method} & \textbf{Time of 10 Iters} & \textbf{Time for One Cycle}\\
    \midrule
    w/o acceleration & 4800 min & unknown\\
    \bestcell{\textbf{w/ lookup table}} & \bestcell{\textbf{3 min}} & \bestcell{\textbf{640 min}}\\
  \bottomrule
\end{tabular}}

  \label{tab:acceleration}
    \centering
    \small
  \captionof{table}{\textsc{\textbf{Efficiency analysis.} All results are about models on PASCAL VOC.}}
  \resizebox{\textwidth}{!}{
  \begin{tabular}{c|c|c|c}
    \toprule
    \textbf{Method} & \textbf{\makecell{Training Time\\of One Cycle}} & \textbf{\makecell{Inference Time\\of One Cycle}} & \textbf{\# of params}\\
    \midrule
    Entropy & 0 & 5 min & 26.3 M\\
    MIAL~\cite{MIAL} & 7 h 13 min & 43 min & 31.9 M\\
    EBAL~\cite{EBAL} & 0 & 181 min & 26.5 M\\
    \bestcell{\textbf{Ours}} & \bestcell{\textbf{(9 h) + 7 h 8 min}} & \bestcell{\textbf{0.5 min}} & \bestcell{\textbf{33.0 M}}\\
  \bottomrule
\end{tabular}}
  \label{tab:efficiency}
\end{minipage}
\vspace{-5pt}
\end{figure*}

\begin{figure}[t]
\vspace{-3pt}
\centering
\includegraphics[width=1.03\columnwidth]{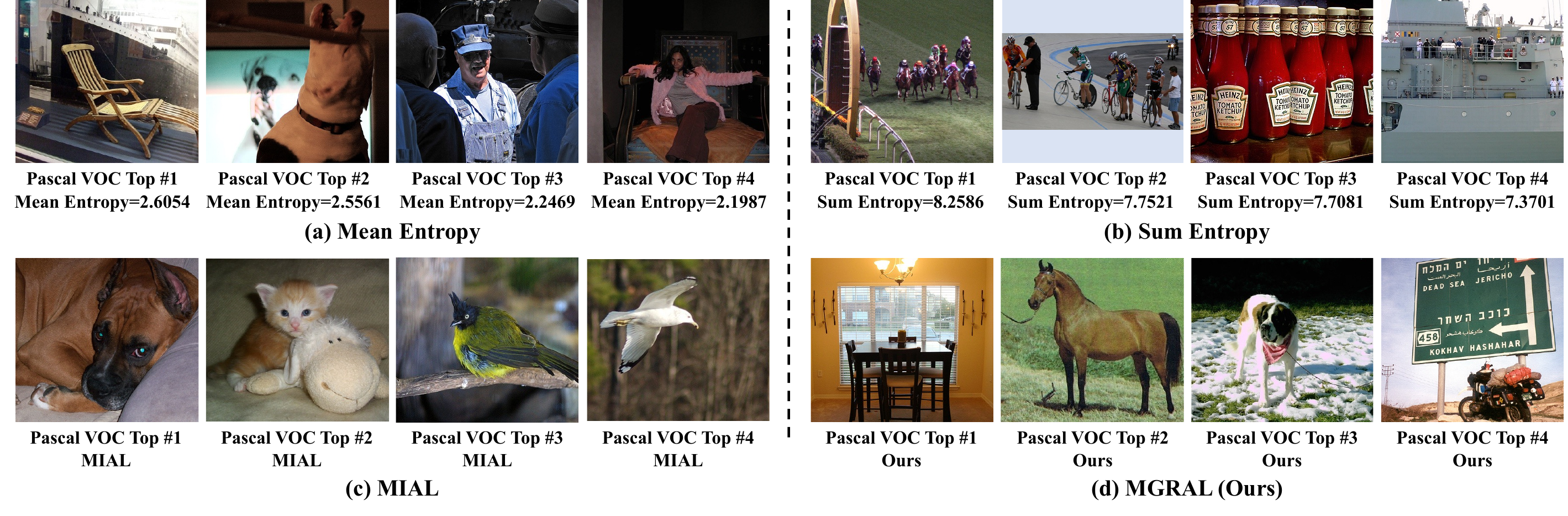}
\vspace{-18pt}
\caption{\textbf{Visualizations of most representative selected samples of different methods during first cycle on Pascal VOC.}}
\vspace{-10pt}
\label{fig:visualization}
\end{figure}

\section{Experiments}
\subsection{Experiment Details}
\textbf{Datasets:} We evaluate \alias on PASCAL VOC~\cite{PASCALVOC} (VOC07+12 trainval: 16,551 images, 20 classes; VOC07 test: 4,952 images) and MS COCO~\cite{COCO} (train2017: 117k images; val2017: 5k images; 80 classes).

\textbf{Active Learning Settings:}
For VOC, we follow~\cite{LL4AL,MIAL,EBAL} with an initial pool of 1,000 labeled images and select 1,000 images per cycle for 10 cycles. For COCO, we start with 2.0\% labeled data, adding 2.0\% per cycle until reaching 10.0\%.

\textbf{Data Sampling Agent Settings (outer loop):}
The controller uses a 257-dimensional vector (256 for image embeddings, 1 for previous selection score) with an LSTM of the same size. We optimize using Adam~\cite{kingma2014adam} (learning rate: $3.5 \times 10^{-4}$). 
The maximum search iterations for each cycle are set to 2,200 for Pascal VOC and 800 for MS COCO, with a baseline decay factor of 0.5 for mAP$_{ref}$.

\textbf{Detector Settings (inner loop):}
On PASCAL VOC, we use SSD~\cite{SSD} detector with ISD-SSD~\cite{jeong2021interpolation}, training for 300 epochs (learning rate: 0.001, reduced to 0.0001 after 240 epochs; batch size: 32). For COCO, we employ RetinaNet~\cite{lin2017focal} detector with SED-SSOD~\cite{guo2022scale} for 180k iterations (batch size: 8; learning rate: 0.02, dropped by 0.1 at the 120k and 160k).

\textbf{Fast Lookup Table:}
For Pascal VOC, we pre-compute 200 experiments per cycle using ISD-SSD. For COCO, we use 30 experiments per cycle with SED-SSOD, totaling 150 records.

\subsection{Quantitative Performance}
We compare \alias with various baselines including random sampling, entropy sampling, Core-set~\cite{coreset}, CDAL~\cite{CDAL}, LL4AL~\cite{LL4AL}, MIAL~\cite{MIAL,wan2023multiple}, and EBAL~\cite{EBAL}. Experiments were conducted on GTX 1080Ti GPUs for PASCAL VOC and Tesla V100 GPUs for MS COCO.

As shown in Fig.\ref{fig:overall_result}, \alias consistently outperforms all methods on PASCAL VOC, demonstrating the effectiveness of our mAP-guided approach. The superior performance can be attributed to our direct optimization of mAP through policy gradient, which naturally aligns the selection strategy with the task's evaluation metric. On MS COCO, \alias shows competitive performance, achieving the steepest performance curve and ultimately surpassing EBAL in the final cycle, though it lags behind EBAL in early cycles. This pattern suggests that while our method excels at fine-grained selection, it may benefit from hierarchical sampling strategies when dealing with large-scale unlabeled pools, potentially using clustering to reduce the search space before applying our selection mechanism.

Our approach's effectiveness stems from two key aspects: first, using mAP as the reward signal provides direct optimization of the detection task's primary metric; second, the reinforcement learning framework enables flexible adaptation to any downstream task by simply changing the reward metric, making our method potentially generalizable beyond detection.

\subsection{Ablation on Visualization Analysis}
Fig.\ref{fig:visualization} compares first-cycle selections across query strategies. Mean entropy sampling (top left) favors exposed or blurred images, potentially hindering initial training. Sum entropy sampling (top right) selects images with multiple instances, but often from redundant categories. MIAL\cite{MIAL} (bottom left) and \alias (bottom right) prefer single, centrally-located objects, providing clearer learning signals. Notably, \alias shows a stronger preference for diverse categories, likely enhancing detector robustness across classes. This diversity-oriented selection explains \alias's superior mAP performance, demonstrating the effectiveness of its mAP-guided approach in creating a balanced and effective initial dataset.

\subsection{Ablation on Efficiency Analysis}
For time complexity, as shown in Tab.\ref{tab:acceleration}, the fast lookup table significantly reduces training time from 4800 to 3 minutes for ten iterations on four GTX 1080Ti GPUs. While this initial training requires more setup time than baselines like entropy sampling, MIAL~\cite{MIAL}, and EBAL~\cite{EBAL}, \alias demonstrates superior efficiency in subsequent cycles through shorter inference times (Tab.\ref{tab:efficiency}). Importantly, this training overhead is acceptable for real-world deployment scenarios, where data collection, selection, and labeling typically occur in offline cycles rather than requiring real-time responses.

In terms of space complexity, our method maintains a competitive memory footprint, with the model requiring $33.0$M parameters compared to EBAL's $26.5$M and MIAL's $31.9$M parameters (Tab.\ref{tab:efficiency}). The additional lookup table introduces minimal overhead, requiring only $12$KB of storage while enabling substantial computational acceleration.

\begin{table}[t]
\centering
\small
\vspace{-2pt}
\caption{\textbf{Ablation on the use of lookup table on VOC.}}
\vspace{-3pt}
\tabcolsep3.5pt
    \resizebox{1.\linewidth}{!}{
  \begin{tabular}{c|c|c|c|c|c|c|c|c|c|c}
    \toprule
    \textbf{Table Size} & \textbf{1000} & \textbf{2000} & \textbf{3000} & \textbf{4000} & \textbf{5000} & \textbf{6000} & \textbf{7000} & \textbf{8000} & \textbf{9000} & \textbf{10000}\\
    \midrule
    \textbf{50 records} & \textbf{56.41} & 62.95 & 67.22 & \textbf{70.33} & 72.12& 73.04& \textbf{74.04}& 74.83& \textbf{75.57}& 75.82\\
    \textbf{100 records} & \textbf{56.41} & 62.91 & 67.56 & 69.82 & \textbf{72.38} & 72.82& 73.90& 74.46& 75.02& 75.57\\
    \bestcell{\textbf{200 records}} & \bestcell{\textbf{56.41}} & \bestcell{\textbf{63.26}} & \bestcell{\textbf{68.09}} & \bestcell{\textbf{70.30}} & \bestcell{72.05} & \bestcell{\textbf{73.16}} & \bestcell{73.96} & \bestcell{\textbf{74.94}} & \bestcell{75.41} & \bestcell{\textbf{75.85}}\\
  \bottomrule
\end{tabular}}
\vspace{-11pt}
  \label{tab:diff-size}
\end{table}

\subsection{Ablation on the Use of Lookup Table}
To investigate the impact and robustness of our lookup table acceleration technique, we conduct experiments with varying numbers of pre-computed records in the fast lookup table. As shown in Tab.\ref{tab:diff-size}, we evaluate the performance using 50, 100, and 200 records across different training stages (1k-10k images) on VOC dataset.

The results demonstrate that as long as the lookup table maintains a reasonable size, our method can achieve superior performance. While 200 records yields slightly better results (75.85 mAP at 10k images), even with just 50 records, the method achieves strong performance (75.82 mAP). This suggests that our acceleration strategy is effective with a modest number of pre-computed records, offering substantial speedup while maintaining performance advantages.

\section{Conclusion}
We presented \alias, a Mean-AP Guided Reinforced Active Learning framework for object detection. By using $\Delta\text{mAP}$ as reward, our method effectively aligns batch selection with detection performance improvement, addressing the non-differentiable nature of this process through policy gradient optimization. The integration of semi-supervised detection and lookup table-based acceleration enables practical deployment while maintaining efficiency. \alias achieves consistent improvements over prior methods on PASCAL VOC and demonstrates competitive performance on MS COCO, establishing a new paradigm that combines performance-driven reinforcement learning with efficient active learning for object detection.

\textbf{Future direction:} Explore early stopping techniques for faster mAP estimation and replacing lookup tables with prediction networks for online reinforcement learning.


\bibliographystyle{IEEEbib}
\bibliography{icme2025references}

\end{document}